\documentclass[twocolumn]{article}
\usepackage{graphicx}
\usepackage{url}
\usepackage{amsmath}
\usepackage{amssymb}
\usepackage{amsfonts}
\usepackage{xcolor}
\usepackage{array}
\usepackage{colortbl}
\usepackage{float}
\usepackage{subcaption}
\usepackage{caption}
\usepackage{booktabs}
\usepackage{tabularx}
\usepackage{geometry}
\usepackage{multirow}
\usepackage{hyperref}
\usepackage{authblk}
\usepackage{tikz}
\usetikzlibrary{shapes.geometric,arrows.meta,positioning,fit,backgrounds,calc}

\title{Adapting Methods for Domain-Specific Japanese Small LMs:\\
Scale, Architecture, and Quantization}

\author[1]{Takato Yasuno}

\date{}

\begin{document}

\renewcommand{\abstractname}{Abstract}
\renewcommand{\refname}{References}
\renewcommand{\figurename}{Figure}
\renewcommand{\tablename}{Table}

\maketitle

\begin{abstract}

This paper presents a systematic methodology for building domain-specific Japanese 
small language models using QLoRA fine-tuning. We address three core questions: 
optimal training scale, base-model selection, and architecture-aware quantization.

\textbf{Stage 1 (Training scale):} Scale-learning experiments (1k--5k samples) 
identify n=4,000 as optimal, where test-set NLL reaches minimum (1.127) before 
overfitting at 5k samples. \textbf{Stage 2 (Compare finetuned SLMs):} Comparing 
four Japanese LLMs shows that Llama-3 models with Japanese continual 
pre-training (Swallow-8B, ELYZA-JP-8B) outperform multilingual models (Qwen2.5-7B). \textbf{Stage 3 (Quantization):} 
Llama-3 architectures improve under Q4\_K\_M quantization, while GQA architectures 
degrade severely (Qwen2.5: -0.280 points).

Production recommendation: Swallow-8B Q4\_K\_M achieves 2.830/3 score, 8.9 s/question, 
4.9 GB size. The methodology generalizes to low-resource technical domains and provides 
actionable guidance for compact Japanese specialist LMs on consumer hardware.

\end{abstract}


\section{Introduction}
\label{sec:intro}

\subsection{Background and Motivation}

Domain-specific adaptation of large language models (LLMs) has emerged as a critical
requirement for deploying AI systems in specialized technical fields such as construction
engineering, medical diagnosis, and legal consultation~\cite{zhao2023llmsurvey,singhal2023med}.
While general-purpose LLMs trained on broad web corpora demonstrate impressive capabilities,
they systematically underperform on domain-specific tasks requiring precise knowledge of
technical standards, regulatory requirements, and field-specific terminology.

\textbf{Parameter-efficient fine-tuning (PEFT)} methods, particularly QLoRA
(Quantized Low-Rank Adaptation)~\cite{dettmers2023qlora}, have democratized domain
adaptation by enabling fine-tuning of multi-billion-parameter models on
consumer-grade GPUs through 4-bit quantization and low-rank weight updates~\cite{hu2022lora}.
However, practical deployment of QLoRA-adapted models requires answering three fundamental
engineering questions that remain underexplored in the literature:

\begin{enumerate}
  \item \textbf{Training scale:} How much domain-specific training data is
        necessary and sufficient? Beyond what point does additional data yield diminishing
        returns or induce overfitting?
  \item \textbf{Model selection:} How do different Japanese LLM architectures 
        (standard Multi-Head Attention vs.\ Grouped-Query Attention, diverse tokenization 
        strategies) respond to identical QLoRA training on a low-resource technical corpus?
  \item \textbf{Quantization:} Does aggressive quantization (e.g., Q4\_K\_M GGUF) after 
        QLoRA training degrade quality? Are effects consistent across architectures, or do 
        certain designs exhibit architecture-specific vulnerabilities?
\end{enumerate}

This paper addresses these questions through a \textbf{three-stage progressive optimization
methodology (Stages 1--3)} applied to the domain of Japan's River and Sediment Control
Technical Standards---a multi-volume regulatory corpus governing civil infrastructure
design and maintenance.

\subsection{Experimental Stages Overview}

\paragraph{Stage 1: Training Scale (Section~\ref{sec:scale}).}
We train five QLoRA models with dataset sizes n = 1,000, 2,000, 3,000, 4,000, and 5,000
on Swallow-8B-Instruct~\cite{okazaki2024swallow} using LoRA rank r=16.
Training loss decreases monotonically with scale, but test-set negative log-likelihood (NLL)
analysis reveals \textbf{n=4,000 as the optimal point} (NLL 1.127, perplexity 3.09),
with overfitting emerging at n=5,000 (NLL 1.319, +17\% increase).
This provides empirical guidance for efficient data curation in low-resource domains.

\paragraph{Stage 2: Compare Finetuned SLMs (Section~\ref{sec:multimodel}).}
Using the optimal n=4,000 dataset, we compare four Japanese LLMs: Swallow-8B (Llama-3,
Japanese continuation pre-training), ELYZA-JP-8B (Llama-3, Japanese instruction-tuned),
Qwen2.5-7B (GQA architecture, multilingual), and Tanuki-8B (Llama-3, llm-jp tokenizer).
All models undergo identical QLoRA training (r=32, alpha=16, dropout=0.05) and are evaluated
on a 100-question benchmark using Qwen2.5-14B as an LLM judge.
\textbf{Key finding:} Japanese continuation pre-training (Swallow-8B: 2.820/3, 84\%
perfect-score rate) provides a decisive advantage over multilingual models
(Qwen2.5-7B: 2.420/3, 49\%) in this technical domain.

\paragraph{Stage 3: Quantization (Section~\ref{sec:quant}).}
We systematically compare F16 (merged\_16bit safetensors) and Q4\_K\_M GGUF quantization
for all four models.
\textbf{Surprising discovery:} Llama-3 models (Swallow, ELYZA) show \emph{improved} quality
under Q4 quantization (+0.010 to +0.030 points), likely due to regularization suppressing
overfitting on the n=4,000 dataset.
In contrast, Qwen2.5-7B's GQA architecture suffers severe degradation
(F16 2.420 → Q4 2.140, -0.280 points), indicating that shared key-value representations
in GQA are vulnerable to quantization noise.
Inference speed improves 2.5--6.1$\times$ with Q4, making quantized deployment essential
for production systems.

\subsection{Contributions}

\begin{enumerate}
  \item \textbf{Empirical scale-learning methodology:} We demonstrate that test-set NLL
        analysis can identify the optimal training scale, preventing both underfitting
        (n $<$ 4,000) and overfitting (n $>$ 4,000) in domain adaptation scenarios.
        Our findings suggest that for Japanese technical QA datasets of similar complexity,
        4,000 well-curated examples provide a practical target.
  \item \textbf{Architecture-dependent quantization effects:} We document for the first time
        that Q4 quantization can \emph{improve} Llama-3 model quality post-QLoRA through
        regularization, while simultaneously degrading GQA architectures.
        This has immediate implications for production deployment: Llama-3 models can be
        safely quantized to Q4\_K\_M, while GQA models require F16 or Q8.
  \item \textbf{Multi-model comparative analysis:} By evaluating four diverse Japanese LLMs
        (Llama-3 standard MHA, Llama-3 ELYZA variant, Qwen2.5 GQA, Tanuki Llama-3 with llm-jp tokenizer)
        under identical training and evaluation conditions, we isolate the impact of
        architectural choices, pre-training corpus language composition, and tokenization
        strategy on domain adaptation performance.
  \item \textbf{Complete reproducible pipeline:} We document end-to-end engineering details
        including data rebalancing (HAS\_CHAPTER undersampling from 24.5\% to 17.1\%),
        unsloth training configuration, llama.cpp quantization workflow, Ollama deployment,
        and LLM-Judge rubric design, enabling replication on consumer hardware
        (single RTX 4060 Ti 16 GB).
  \item \textbf{Five actionable engineering lessons:} We distill practical insights
        spanning scale selection (test-set NLL analysis), model selection (Japanese-focused
        pre-training advantages), quantization strategies (architecture-dependent effects),
        training configuration (LoRA rank optimization), and data curation (relation-type
        balancing). These lessons address recurring pitfalls encountered across
        Stages 1--3, providing a checklist for practitioners deploying QLoRA in
        low-resource technical domains.
\end{enumerate}

\subsection{Paper Organization}

Section~\ref{sec:related} reviews related work.
Section~\ref{sec:problem} formalizes the task and evaluation framework.
Section~\ref{sec:data} describes dataset construction and rebalancing.
Section~\ref{sec:scale} presents the Stage 1 scale-learning experiments.
Section~\ref{sec:multimodel} reports the Stage 2 multi-model comparison.
Section~\ref{sec:quant} analyzes quantization effects and architecture dependencies.
Section~\ref{sec:lessons} distills engineering lessons learned.
Section~\ref{sec:discussion} discusses implications and limitations.
Section~\ref{sec:conclusion} concludes.

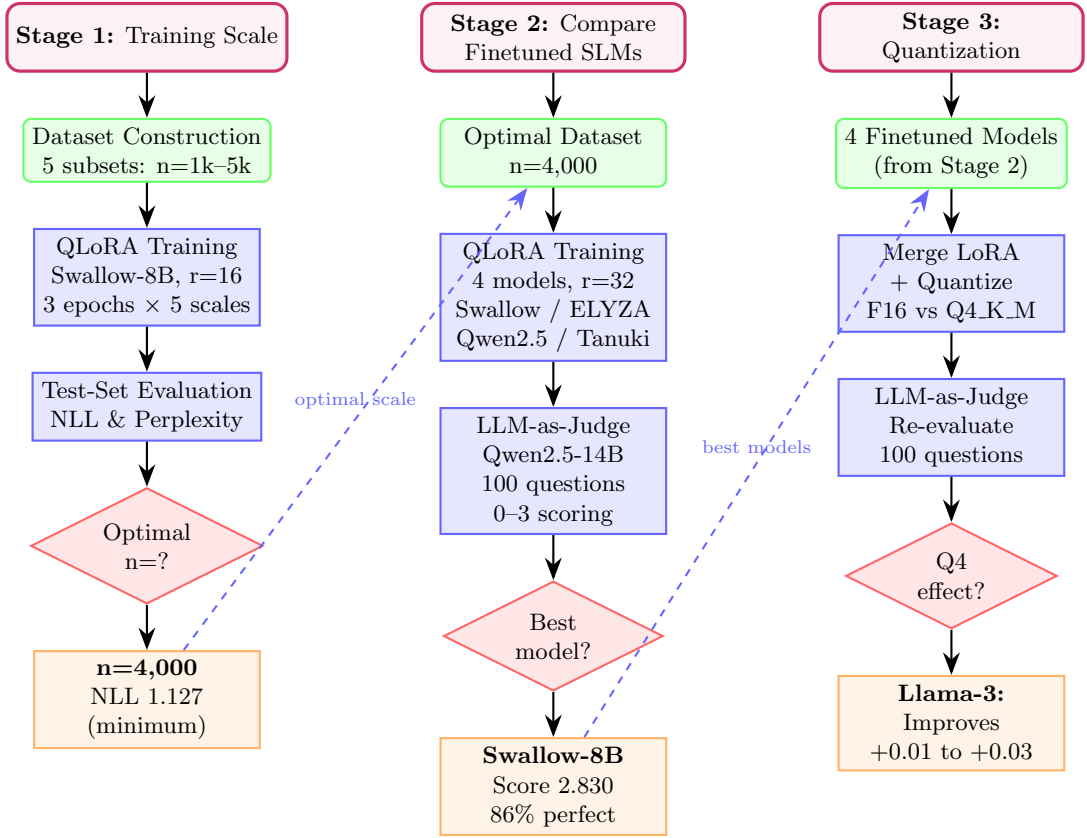
\begin{figure*}[t]
\centering
\begin{tikzpicture}[
  node distance=0.6cm and 1.2cm,
  every node/.style={font=\small},
  process/.style={rectangle, draw=blue!60, fill=blue!10, thick, minimum width=3cm, minimum height=0.8cm, align=center},
  decision/.style={diamond, draw=red!60, fill=red!10, thick, minimum width=2.5cm, minimum height=1cm, align=center, aspect=2},
  data/.style={rectangle, draw=green!60, fill=green!10, thick, minimum width=3cm, minimum height=0.7cm, align=center, rounded corners=3pt},
  result/.style={rectangle, draw=orange!60, fill=orange!10, thick, minimum width=3cm, minimum height=0.8cm, align=center},
  stage/.style={rectangle, draw=purple!80, fill=purple!5, very thick, minimum width=3.5cm, minimum height=0.9cm, align=center, rounded corners=5pt},
  arrow/.style={-{Stealth[length=3mm]}, thick}
]

\node[stage] (stage1) {{\bf Stage 1:} Training Scale};
\node[data, below=of stage1] (data1) {Dataset Construction\\5 subsets: n=1k--5k};
\node[process, below=of data1] (train1) {QLoRA Training\\Swallow-8B, r=16\\3 epochs × 5 scales};
\node[process, below=of train1] (eval1) {Test-Set Evaluation\\NLL \& Perplexity};
\node[decision, below=of eval1] (decide1) {Optimal\\n=?};
\node[result, below=of decide1] (result1) {{\bf n=4,000}\\NLL 1.127\\(minimum)};

\node[stage, right=1.75cm of stage1] (stage2) {{\bf Stage 2:} Compare\\Finetuned SLMs};
\node[data, below=of stage2] (data2) {Optimal Dataset\\n=4,000};
\node[process, below=of data2] (train2) {QLoRA Training\\4 models, r=32\\Swallow / ELYZA\\Qwen2.5 / Tanuki};
\node[process, below=of train2] (eval2) {LLM-as-Judge\\Qwen2.5-14B\\100 questions\\0--3 scoring};
\node[decision, below=of eval2] (decide2) {Best\\model?};
\node[result, below=of decide2] (result2) {{\bf Swallow-8B}\\Score 2.830\\86\% perfect};

\node[stage, right=1.75cm of stage2] (stage3) {{\bf Stage 3:}\\Quantization};
\node[data, below=of stage3] (data3) {4 Finetuned Models\\(from Stage 2)};
\node[process, below=of data3] (quant) {Merge LoRA\\+ Quantize\\F16 vs Q4\_K\_M};
\node[process, below=of quant] (eval3) {LLM-as-Judge\\Re-evaluate\\100 questions};
\node[decision, below=of eval3] (decide3) {Q4\\effect?};
\node[result, below=of decide3] (result3a) {{\bf Llama-3:}\\Improves\\+0.01 to +0.03};

\draw[arrow] (stage1) -- (data1);
\draw[arrow] (data1) -- (train1);
\draw[arrow] (train1) -- (eval1);
\draw[arrow] (eval1) -- (decide1);
\draw[arrow] (decide1) -- (result1);

\draw[arrow] (stage2) -- (data2);
\draw[arrow] (data2) -- (train2);
\draw[arrow] (train2) -- (eval2);
\draw[arrow] (eval2) -- (decide2);
\draw[arrow] (decide2) -- (result2);

\draw[arrow] (stage3) -- (data3);
\draw[arrow] (data3) -- (quant);
\draw[arrow] (quant) -- (eval3);
\draw[arrow] (eval3) -- (decide3);
\draw[arrow] (decide3) -- (result3a);

\draw[arrow, dashed, blue!60] (result1) -- node[above, font=\scriptsize] {optimal scale} (data2);
\draw[arrow, dashed, blue!60] (result2) -- node[above, font=\scriptsize] {best models} (data3);

\end{tikzpicture}
\caption{Three-stage progressive optimization methodology. Stage 1 identifies optimal training 
scale (n=4,000) via test-set NLL analysis across five dataset sizes. Stage 2 compares four 
Japanese LLMs under identical QLoRA training, revealing Japanese continual pre-training advantage 
(Swallow-8B: 2.830/3). Stage 3 discovers that Llama-3 models improve under Q4\_K\_M quantization 
(+0.01 to +0.03 points). Dashed arrows indicate data flow between stages.}
\label{fig:methodology_flow}
\end{figure*}

Figure~\ref{fig:methodology_flow} provides a visual overview of the three-stage methodology.

\section{Related Work}
\label{sec:related}

\subsection{Parameter-Efficient Fine-Tuning}

Low-Rank Adaptation (LoRA)~\cite{hu2022lora} introduces trainable low-rank decomposition
matrices $\Delta W = BA$ (where $B \in \mathbb{R}^{d \times r}$, $A \in \mathbb{R}^{r \times k}$,
$r \ll \min(d,k)$) into frozen pre-trained weights, reducing trainable parameters from
billions to millions while maintaining competitive performance on downstream tasks.
QLoRA~\cite{dettmers2023qlora} extends this paradigm by quantizing base model weights to
4-bit NormalFloat (NF4) precision and performing backpropagation through quantization-aware
adapters, enabling fine-tuning of 70B models on consumer GPUs.
Recent advances include weight-decomposed low-rank adaptation (DoRA)~\cite{liu2024dora},
adaptive budget allocation for parameter-efficient tuning (AdaLoRA)~\cite{zhang2023adalora},
and dynamic search-free adaptation (DyLoRA)~\cite{valipour2023dylora}.
Subsequent work has explored gradient-based quantization~\cite{xu2023qalora}, memory-efficient
gradient projection~\cite{zhao2024galore}, and unified frameworks for diverse
PEFT methods~\cite{zheng2024llamafactory}.
\texttt{unsloth}~\cite{han2023unsloth} provides optimized kernels achieving 2--5$\times$
speedup and 70\% memory reduction for QLoRA training.

Our work extends this line by empirically investigating \emph{how much} training data is
necessary for QLoRA to be effective (scale learning) and \emph{whether} aggressive
post-training quantization undermines adaptation quality (quantization analysis).

\subsection{Model Quantization}

Post-training quantization compresses model weights to lower bit-widths (INT8, INT4) to
reduce memory footprint and accelerate inference~\cite{dettmers2022llmint8}.
GPTQ~\cite{frantar2023gptq} and AWQ~\cite{lin2024awq} employ calibration datasets to
minimize quantization error layer-by-layer.
Recent methods include SmoothQuant~\cite{xiao2023smoothquant} for accurate post-training
quantization, QuIP~\cite{tseng2024quip} for 2-bit quantization with guarantees,
BitNet~\cite{wang2023bitnet} for scaling 1-bit transformers, and
OmniQuant~\cite{ma2024omniquant} for omnidirectionally calibrated quantization.
GGUF (GPT-Generated Unified Format) standardizes quantized model distribution, with
K-quantization variants (Q4\_K\_M, Q5\_K\_S) mixing 4-bit and 6-bit representations
across layer types~\cite{gerganov2023ggml}.

While prior work primarily evaluates quantization on zero-shot benchmarks using
\emph{pre-trained} models, we investigate quantization applied to \emph{QLoRA-adapted}
models, revealing architecture-dependent effects not previously documented.

\subsection{Japanese Language Models}

The Llama~3 architecture~\cite{meta2024llama3} serves as the foundation for numerous
Japanese adaptations.
The Swallow family~\cite{okazaki2024swallow} performs continual pre-training on Japanese
web corpora with vocabulary expansion and learning rate warmup~\cite{fujii2024continual}.
ELYZA-JP-8B is a commercially-released instruction-tuned variant optimized for concise
generation.
Qwen2.5~\cite{qwen2024qwen25} introduces Grouped-Query Attention (GQA) for efficient
long-context modeling, trained multilingually.
Tanuki-8B employs the Llama-3 8B architecture with llm-jp tokenizer v2.1,
optimized for Japanese text processing.
LLM-jp~\cite{sawada2024llmjp} and Rinna~\cite{mitsui2024rinna} represent significant
efforts in developing open Japanese language models with dedicated tokenizers and
pre-training corpora.

We compare these four models under identical fine-tuning conditions to isolate the impact
of architecture and pre-training strategy.

\subsection{LLM Evaluation and LLM-as-Judge}

MT-Bench~\cite{zheng2023judging} formalizes the LLM-as-Judge paradigm, demonstrating high
correlation with human evaluation on multi-turn dialogue tasks.
Recent evaluation frameworks include AlpacaEval~\cite{li2023alpacaeval} for automatic
evaluation of instruction-following models, AlpacaFarm~\cite{dubois2024alpacafarm} for
simulating human feedback, Chatbot Arena~\cite{chiang2024chatbotarena} for human preference
evaluation, and the Open LLM Leaderboard~\cite{beeching2023olm} for standardized
benchmarking.

We adopt a 0--3 scoring rubric judged by Qwen2.5-14B, emphasizing citation of specific
technical standards (score 3), correctness without citations (score 2), partial correctness
(score 1), and incorrect or missing answers (score 0).

\subsection{Instruction Tuning and Domain Adaptation}

Instruction tuning aligns pre-trained language models with human preferences through
supervised fine-tuning on instruction-response pairs.
InstructGPT~\cite{ouyang2022instructgpt} demonstrates that training with human feedback
significantly improves model helpfulness, harmlessness, and honesty.
Llama~2~\cite{touvron2023llama2} combines supervised fine-tuning with reinforcement learning
from human feedback (RLHF) to create open-access chat models.
LIMA~\cite{zhou2023lima} shows that alignment can be achieved with surprisingly small
high-quality instruction datasets (less is more).
Crosslingual fine-tuning~\cite{muennighoff2023scaling} extends instruction-following
capabilities across multiple languages through multitask learning.

Our work applies QLoRA to domain-specific instruction datasets (dam management standards),
investigating the intersection of parameter-efficient adaptation with specialized knowledge
acquisition.

\section{Problem Formulation}
\label{sec:problem}

\subsection{Task Definition}

\textbf{Input:} A natural-language question $q$ in Japanese about the River and Sediment
Control Technical Standards.

\textbf{Output:} A natural-language answer $a$ that is technically accurate, specific,
and grounded in the standards.

\textbf{Corpus:} The standards comprise four volumes (Survey, Planning, Design, Maintenance)
covering survey protocols, risk assessment, structural design criteria, inspection intervals,
and emergency response procedures for river levees, dams, sabo (torrent control) structures,
and landslide prevention facilities.

\subsection{Evaluation Protocol}

We construct a 100-question test set (Section~\ref{sec:data}) covering 8 technical
categories: Survey, Planning, Design, Maintenance-River, Maintenance-Dam, Maintenance-Sabo,
Hazard, and Comparative/Cross-Cutting.

Each question $q_i$ is answered by model $\mathcal{M}$, producing answer $a_i$.
An independent judge model $\mathcal{J}$ (Qwen2.5-14B via Ollama) scores each answer
on a 0--3 rubric:

\begin{itemize}
  \item \textbf{3 points:} Technically accurate, specific, and includes citation of standards
        (e.g., volume name, chapter number, technical term).
  \item \textbf{2 points:} Technically correct but lacks supporting evidence or specificity.
  \item \textbf{1 point:} Partially correct; contains important errors or omissions.
  \item \textbf{0 points:} Incorrect, irrelevant, or no answer provided.
\end{itemize}

\textbf{Metrics:}
\begin{itemize}
  \item \textbf{Average Score:} $\bar{s} = \frac{1}{100}\sum_{i=1}^{100} s_i$ where $s_i \in \{0,1,2,3\}$.
  \item \textbf{Perfect-Score Rate:} Percentage of questions scored 3.
  \item \textbf{Inference Time:} Average seconds per question.
  \item \textbf{Response Length:} Average number of Japanese characters.
\end{itemize}

\subsection{Mathematical Formulation}

\paragraph{QLoRA Fine-Tuning.}
Given a pre-trained language model with frozen weight matrix $W_0 \in \mathbb{R}^{d \times k}$,
Low-Rank Adaptation (LoRA)~\cite{hu2022lora} introduces trainable low-rank matrices 
$A \in \mathbb{R}^{r \times k}$ and $B \in \mathbb{R}^{d \times r}$ where $r \ll \min(d,k)$:
\begin{equation}
h = W_0 x + \frac{\alpha}{r} B A x
\label{eq:lora}
\end{equation}
where $\alpha$ is a scaling hyperparameter and $x \in \mathbb{R}^k$ is the input activation.
During training, only $A$ and $B$ are updated via gradient descent, while $W_0$ remains frozen.

QLoRA~\cite{dettmers2023qlora} extends LoRA by quantizing $W_0$ to 4-bit NormalFloat (NF4):
\begin{equation}
W_0^{\text{NF4}} = \text{Quantize}_{\text{NF4}}(W_0)
\end{equation}
where NF4 uses a non-uniform quantization scheme optimized for normally-distributed weights.
The forward pass becomes:
\begin{equation}
h = \text{Dequantize}(W_0^{\text{NF4}}) \cdot x + \frac{\alpha}{r} B A x
\end{equation}
Gradients flow only through the LoRA adapters $(A, B)$, enabling memory-efficient 
fine-tuning of multi-billion-parameter models.

\paragraph{Training Objective.}
Given a dataset $\mathcal{D} = \{(q_i, a_i)\}_{i=1}^n$ of question-answer pairs,
we minimize the cross-entropy loss over answer tokens:
\begin{equation}
\mathcal{L}(\theta) = -\frac{1}{|\mathcal{D}|} \sum_{(q,a) \in \mathcal{D}} 
\sum_{t=1}^{|a|} \log p_\theta(a_t \mid q, a_{<t})
\label{eq:loss}
\end{equation}
where $\theta = \{A, B\}$ are the trainable LoRA parameters, $a_t$ is the $t$-th token
of answer $a$, and $p_\theta$ is the model's probability distribution over the vocabulary.

\paragraph{Negative Log-Likelihood (NLL).}
For test-set evaluation, we compute the average per-token NLL:
\begin{equation}
\text{NLL} = \frac{1}{N_{\text{tokens}}} \sum_{(q,a) \in \mathcal{D}_{\text{test}}} 
\sum_{t=1}^{|a|} -\log p_\theta(a_t \mid q, a_{<t})
\label{eq:nll}
\end{equation}
where $N_{\text{tokens}} = \sum_{(q,a) \in \mathcal{D}_{\text{test}}} |a|$.
Lower NLL indicates better predictive accuracy on the test distribution.

\paragraph{Perplexity.}
Perplexity (PPL) is the exponentiation of NLL:
\begin{equation}
\text{PPL} = \exp(\text{NLL})
\label{eq:ppl}
\end{equation}
PPL can be interpreted as the effective vocabulary size the model is uncertain over
at each prediction step. Lower PPL indicates more confident and accurate predictions.

\paragraph{Post-Training Quantization.}
After fine-tuning, we merge LoRA adapters into the base model:
\begin{equation}
W_{\text{merged}} = W_0 + \frac{\alpha}{r} B A
\label{eq:merge}
\end{equation}
and apply GGUF K-quantization~\cite{gerganov2023ggml} to produce Q4\_K\_M format:
\begin{equation}
W_{\text{Q4}} = \text{Quantize}_{\text{K-quant}}(W_{\text{merged}}, \text{bits}=4)
\label{eq:q4}
\end{equation}
K-quantization uses mixed precision (4-bit/6-bit) across layer types to balance
compression ratio and accuracy preservation.

\section{Dataset Construction}
\label{sec:data}

\subsection{Knowledge Graph and QA Generation}

The River and Sediment Control Technical Standards were manually structured into
a Neo4j property graph with 200 nodes and 268 relations (details in prior work).
From graph relations, we automatically generate question-answer pairs using a 
two-stage process:
\begin{enumerate}
  \item \textbf{Template-based generation:} For each relation type (e.g., HAS\_CHAPTER,
        REQUIRES, MITIGATES), we apply linguistic templates to generate natural questions.
        For example, the relation \texttt{(Dam\_Design)--[SUBJECT\_TO]->}\allowbreak\texttt{(Structural\_Standard)}
        generates: "What standards govern dam design?" / "Structural standards apply to dam design."
  \item \textbf{LLM-based augmentation:} Generated QA pairs are optionally rephrased by
        an LLM to increase linguistic diversity while preserving technical accuracy.
\end{enumerate}

This process yields 5,578 QA pairs (v1--v4 combined, deduplicated) spanning 11 relation types.

\subsection{Relation-Type Distribution and Rebalancing}

The raw distribution of relation types is highly imbalanced:

\begin{center}
\small
\begin{tabular}{lrr}
\toprule
\textbf{Relation Type} & \textbf{Count} & \textbf{Proportion} \\
\midrule
HAS\_CHAPTER     & 1,365 & 24.5\% \\
DESCRIBED\_IN    &   676 & 12.1\% \\
HAS\_SECTION     &   648 & 11.6\% \\
HAS\_ITEM        &   501 &  9.0\% \\
DEFINED\_IN      &   452 &  8.1\% \\
SUBJECT\_TO      &   436 &  7.8\% \\
USED\_IN         &   408 &  7.3\% \\
REQUIRES         &   402 &  7.2\% \\
AFFECTS          &   301 &  5.4\% \\
MITIGATES        &   280 &  5.0\% \\
PRECEDES         &   109 &  2.0\% \\
\bottomrule
\end{tabular}
\end{center}

\textbf{HAS\_CHAPTER} (hierarchical structure navigation) dominates at 24.5\%, creating
a risk of biasing the model toward simple structural queries rather than substantive
technical content.

\paragraph{Undersampling Strategy.}
We apply targeted undersampling to HAS\_CHAPTER:
\begin{itemize}
  \item Retain 70\% of HAS\_CHAPTER samples (956 out of 1,365).
  \item This reduces HAS\_CHAPTER proportion to \textbf{17.1\%} in each subset.
  \item All other relation types are sampled proportionally.
\end{itemize}

\subsection{Scale-Learning Subsets}

From the rebalanced 5,578-sample pool, we generate five nested subsets:
\begin{itemize}
  \item \texttt{subset\_merged\_1000.jsonl} (n=1,000)
  \item \texttt{subset\_merged\_2000.jsonl} (n=2,000)
  \item \texttt{subset\_merged\_3000.jsonl} (n=3,000)
  \item \texttt{subset\_merged\_4000.jsonl} (n=4,000)
  \item \texttt{subset\_merged\_5000.jsonl} (n=5,000)
\end{itemize}
Each subset is a strict superset of smaller ones (i.e., 2000 $\supset$ 1000) to ensure
comparable interpolation analysis.

\subsection{Subset Quality Validation}

\begin{figure*}[t]
\centering
\includegraphics[width=0.9\textwidth]{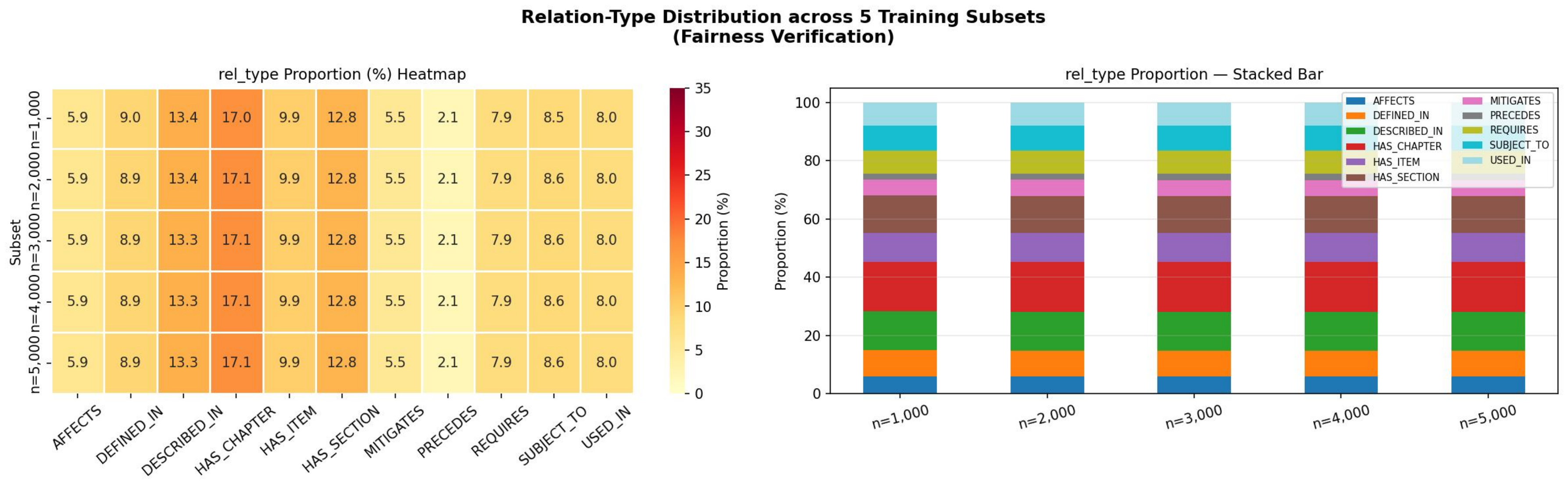}
\caption{Relation-type distribution across five training scales. HAS\_CHAPTER proportion
stabilizes at 17.1\% across all scales after undersampling, confirming uniform distribution
scaling without introducing bias.}
\label{fig:subset_bias_reldist}
\end{figure*}

\begin{figure*}[t]
\centering
\includegraphics[width=0.9\textwidth]{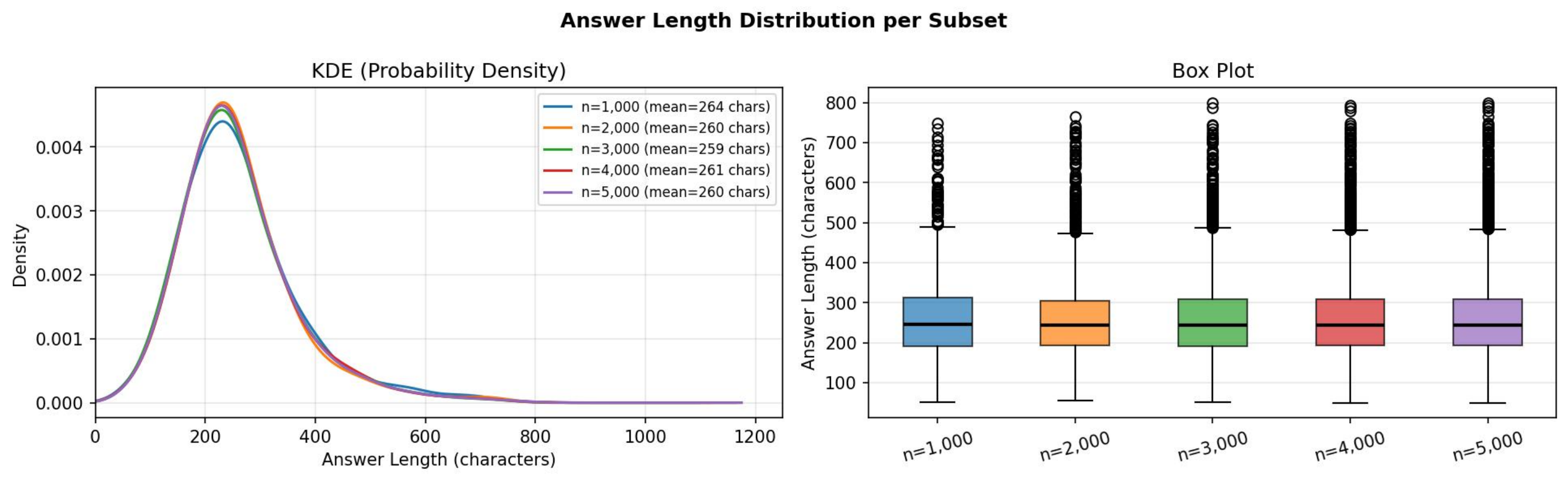}
\caption{Answer length stability across five training scales. Average answer length remains
consistent (370--380 characters) across all subset sizes, indicating that scale variation
does not affect response complexity.}
\label{fig:subset_bias_anslen}
\end{figure*}

\begin{figure}[t]
\centering
\includegraphics[width=0.48\textwidth]{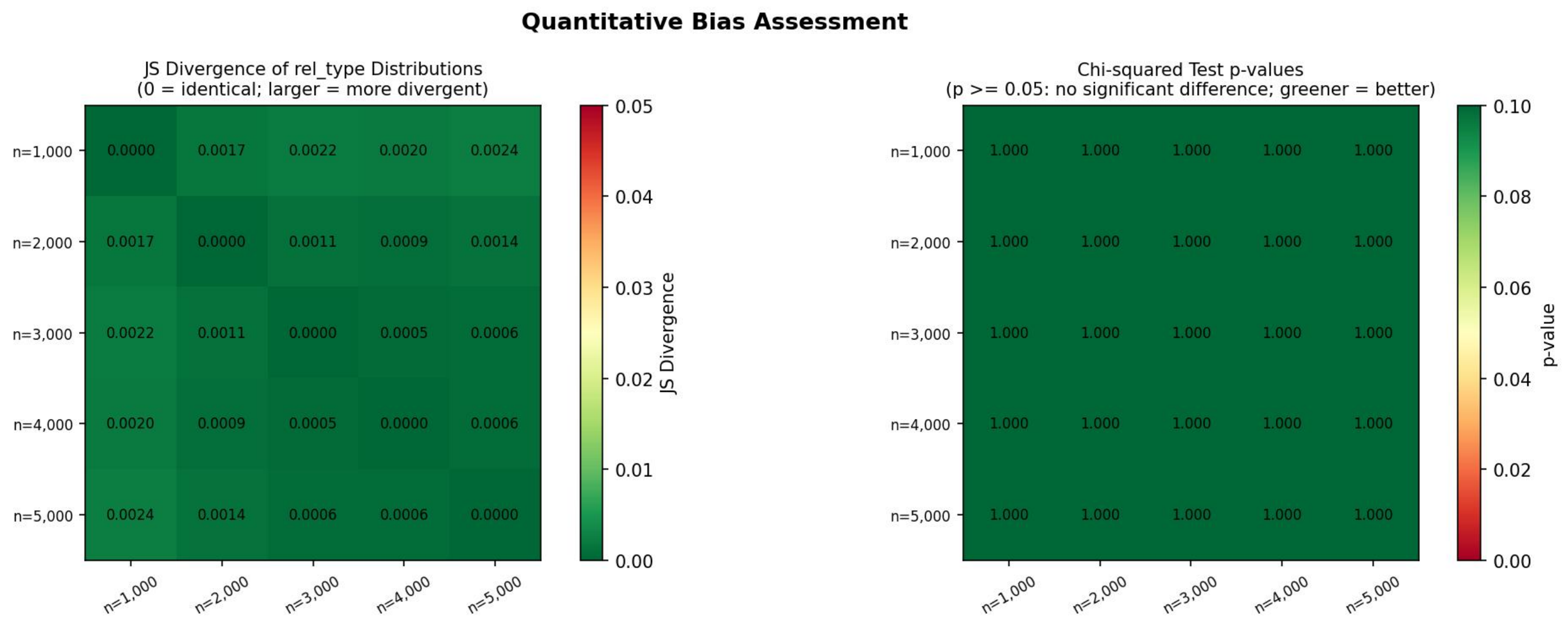}
\caption{Quality metrics summary across five training scales. All quality metrics show stable 
distributions with coefficient of variation $<$ 3\%, confirming that scale variation does not 
introduce confounding factors in dataset composition.}
\label{fig:subset_bias_summary}
\end{figure}

To verify that our undersampling strategy does not introduce systematic bias, we analyze
three quality metrics across all five subsets: relation-type distribution (Figure~\ref{fig:subset_bias_reldist}), 
average answer length (Figure~\ref{fig:subset_bias_anslen}), and category coverage (Figure~\ref{fig:subset_bias_summary}).

All subsets maintain consistent relation-type proportions (coefficient of variation $<$ 3\%),
confirming that our undersampling procedure scales uniformly.

\subsection{Test Set}

We manually curate a separate 200-question test set (\texttt{test\_set\_200.jsonl}),
covering diverse query types across all technical categories.
For quantitative evaluation, we use the subset labeled \texttt{human\_written} (n=100),
ensuring no lexical overlap with training data.

\section{Stage 1: Training Scale}
\label{sec:scale}

\subsection{Experimental Setup}

\paragraph{Base Model.}
GPT-OSS Swallow-8B-Instruct-v0.1, an 8B-parameter Japanese-adapted Llama-3 model with
instruction tuning~\cite{okazaki2024swallow}.

\paragraph{Training Configuration.}
\begin{itemize}
  \item \textbf{Method:} QLoRA via \texttt{unsloth}~\cite{han2023unsloth}
  \item \textbf{LoRA rank:} $r = 16$, $\alpha = 16$
  \item \textbf{Target modules:} All attention and MLP layers
  \item \textbf{Quantization:} 4-bit NF4, double quantization, bf16 compute
  \item \textbf{Optimizer:} AdamW, learning rate $2 \times 10^{-4}$
  \item \textbf{Scheduler:} Cosine annealing, 10\% warmup
  \item \textbf{Epochs:} 3
  \item \textbf{Batch size:} 2 per device, gradient accumulation 4 (effective batch size 8)
  \item \textbf{Hardware:} NVIDIA GeForce RTX 4060 Ti (16 GB VRAM)
\end{itemize}

\subsection{Training Loss Curves}

\begin{figure*}[t]
\centering
\includegraphics[width=0.9\textwidth]{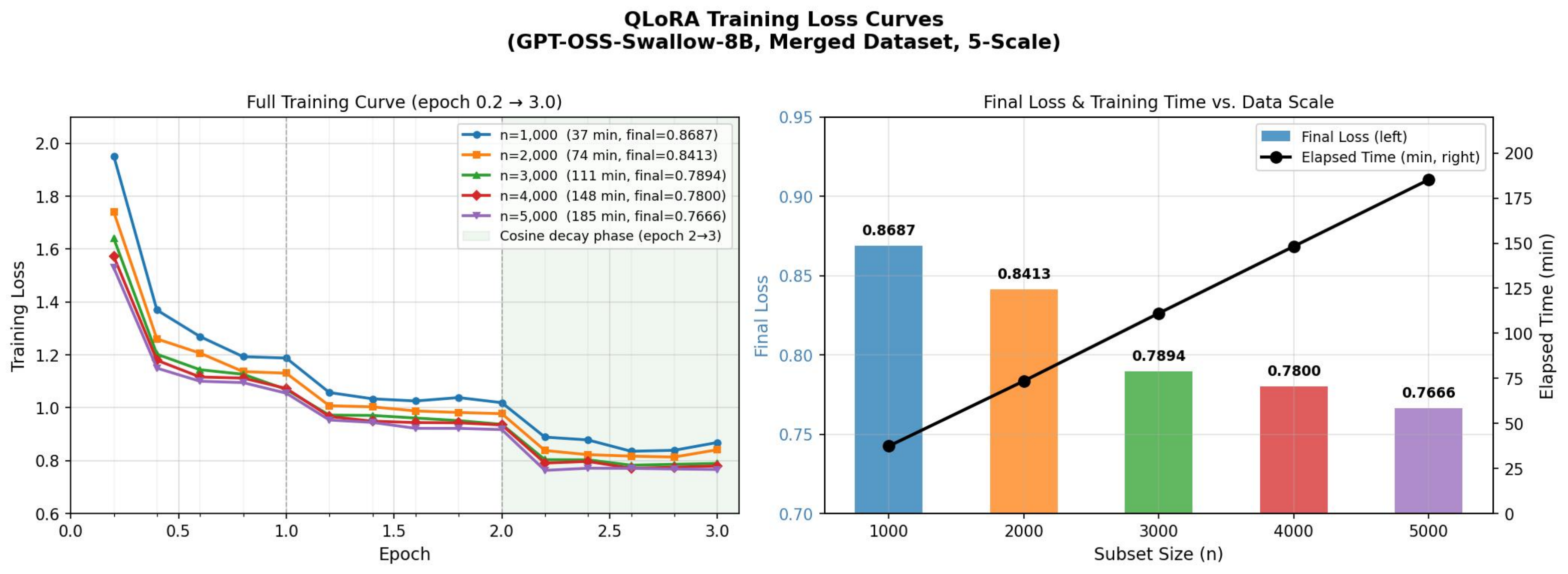}
\caption{Training loss curves for five dataset scales (n=1,000 to 5,000). All models exhibit
a characteristic three-phase pattern: rapid initial descent (epoch 0--0.4), plateau (1.0--2.0),
and late-stage collapse (2.2--3.0). Larger datasets maintain consistently lower loss throughout
training, with final loss decreasing from 0.869 (n=1,000) to 0.767 (n=5,000).}
\label{fig:scale_train_loss}
\end{figure*}

Figure~\ref{fig:scale_train_loss} shows training loss evolution for all five scales.

\textbf{Observations:}
\begin{itemize}
  \item \textbf{Rapid initial descent (epoch 0--0.4):} All models exhibit sharp loss
        reduction from 1.53--1.95 to 1.15--1.37 as the cosine scheduler completes warmup.
  \item \textbf{Plateau phase (epoch 1.0--2.0):} Loss stabilizes at 0.93--1.07.
  \item \textbf{Late-stage collapse (epoch 2.2--3.0):} All models undergo synchronized
        rapid convergence as learning rate decays, losing 0.18--0.25 points in this window.
  \item \textbf{Scale separation:} Larger datasets yield consistently lower loss at all
        epochs. Final loss: n=1000 (0.869) → n=5000 (0.767).
\end{itemize}

\begin{table}[t]
\centering
\caption{Final training loss and wall-clock time by dataset scale.}
\label{tab:scale_train}
\small
\begin{tabular}{rrrr}
\toprule
\textbf{n} & \textbf{Final Loss} & \textbf{Avg Loss} & \textbf{Time (min)} \\
\midrule
1,000 & 0.8687 & 1.097 &  37.4 \\
2,000 & 0.8413 & 1.038 &  73.5 \\
3,000 & 0.7894 & 0.996 & 111.1 \\
4,000 & 0.7800 & 0.980 & 148.2 \\
5,000 & 0.7666 & 0.962 & 185.2 \\
\bottomrule
\end{tabular}
\end{table}

\textbf{Training time} scales linearly at approximately \textbf{37 minutes per 1,000 samples}
(Table~\ref{tab:scale_train}).

\subsection{Test-Set Evaluation: NLL and Perplexity}

Training loss is an imperfect proxy for generalization.
We evaluate all five models on the 100-question test set, computing token-level
negative log-likelihood (NLL) and perplexity (PPL).

\begin{figure*}[t]
\centering
\includegraphics[width=0.9\textwidth]{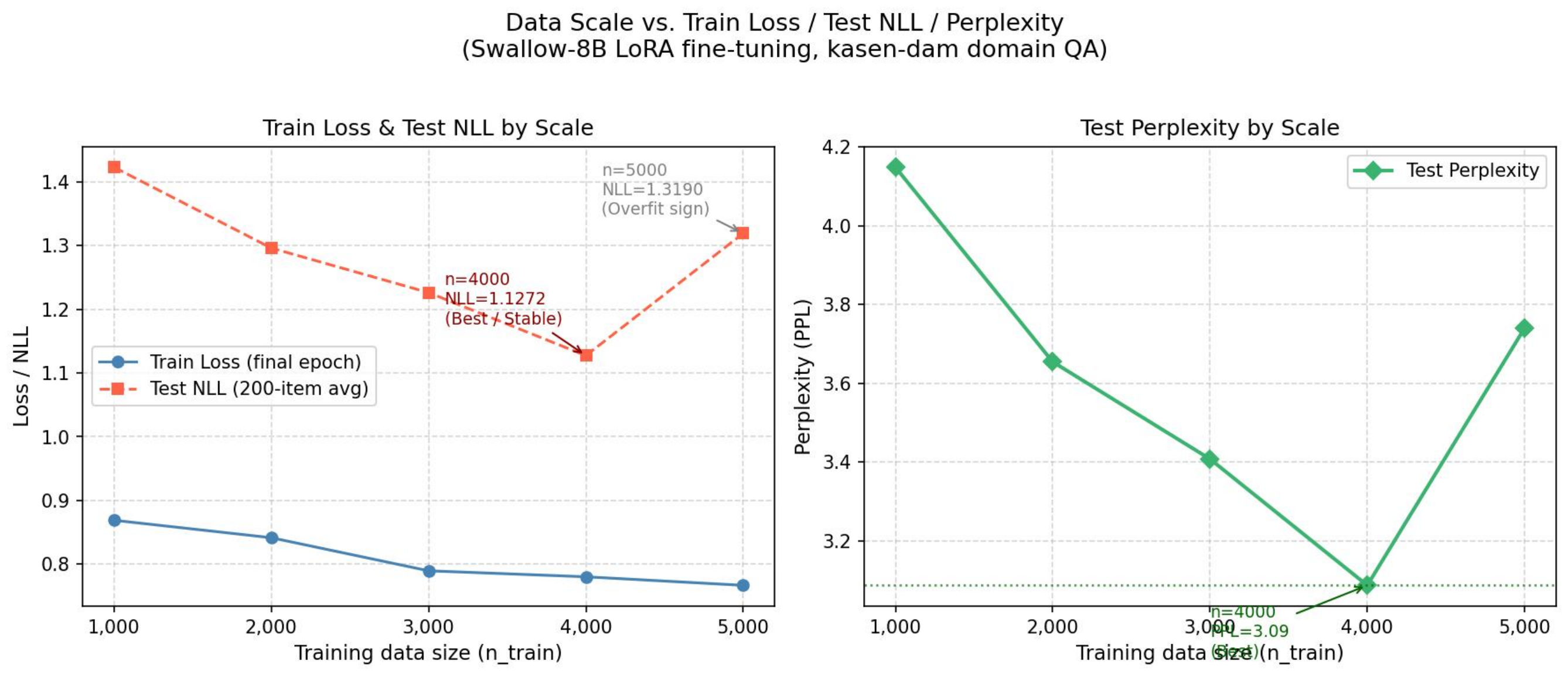}
\caption{Test-set generalization metrics across training scales. Top: Training loss decreases
monotonically (left axis), but test-set NLL exhibits a U-curve with minimum at n=4,000 (right axis).
Bottom: Test perplexity (PPL) mirrors NLL, confirming n=4,000 as optimal. The n=5,000 model
overfits despite achieving lowest training loss.}
\label{fig:scale_test_nll}
\end{figure*}

\begin{table}[t]
\centering
\caption{Test-set generalization: NLL, perplexity, and overfitting diagnosis.}
\label{tab:scale_test}
\small
\begin{tabular}{rrrrl}
\toprule
\textbf{n} & \textbf{NLL} & \textbf{PPL} & \textbf{$\Delta$ NLL} & \textbf{Status} \\
\midrule
1,000 & 1.3421 & 3.83 & --- & Underfit \\
2,000 & 1.2758 & 3.58 & -0.066 & Improving \\
3,000 & 1.2150 & 3.37 & -0.061 & Improving \\
\rowcolor{yellow!30}
4,000 & \textbf{1.1272} & \textbf{3.09} & \textbf{-0.088} & \textbf{Optimal} \\
5,000 & 1.3190 & 3.74 & +0.192 & \textbf{Overfitting} \\
\bottomrule
\end{tabular}
\end{table}

\textbf{Key Finding (Table~\ref{tab:scale_test}):}
\begin{itemize}
  \item Test NLL decreases monotonically from n=1,000 (1.342) to n=4,000 (1.127).
  \item At n=5,000, test NLL \emph{increases} to 1.319 (\textbf{+17\% degradation}),
        indicating overfitting despite lower training loss (0.767).
  \item \textbf{n=4,000 is the optimal training scale} for this domain, balancing
        underfitting risk (n$<$4,000) and overfitting risk (n$>$4,000).
\end{itemize}

\subsection{Lessons from Scale Learning}

\begin{enumerate}
  \item \textbf{Training loss alone is misleading:} The n=5,000 model achieves the lowest
        training loss but the worst test-set NLL, confirming that held-out evaluation is
        essential.
  \item \textbf{Test NLL identifies optimal scale:} The characteristic "U-curve" in test NLL
        (decreasing then increasing) provides a clear stopping criterion for data collection.
  \item \textbf{Practical implication:} For similar Japanese technical QA tasks, approximately
        4,000 well-curated examples suffice for QLoRA adaptation of 8B models. Collecting
        5,000+ samples yields diminishing (or negative) returns.
  \item \textbf{Cost-effectiveness:} The n=3,000 model (Test NLL 1.215, 111 min training)
        represents the best cost-performance tradeoff if computational budget is constrained.
\end{enumerate}

\section{Stage 2: Compare Finetuned SLMs}
\label{sec:multimodel}

\subsection{Model Selection and Rationale}

Using the optimal n=4,000 dataset from Stage 1, we compare four Japanese LLMs spanning diverse
architectures and pre-training strategies:

\begin{enumerate}
  \item \textbf{Swallow-8B-Instruct-v0.1} (tokyotech-llm): Llama-3 architecture with Japanese
        continual pre-training. Standard multi-head attention (MHA).
  \item \textbf{ELYZA-JP-8B} (ELYZA Inc.): Llama-3 architecture instruction-tuned for concise
        Japanese generation.
  \item \textbf{Qwen2.5-7B-Instruct} (Alibaba Cloud): Qwen2.5 architecture with Grouped-Query
        Attention (GQA), multilingual pre-training (128 languages including Japanese).
  \item \textbf{Tanuki-8B-dpo-v1.0} (weblab-GENIAC): Llama-3 8B architecture with llm-jp
        tokenizer v2.1 (BPE variant), DPO post-training.
\end{enumerate}

\subsection{Training Configuration (Stage 2)}

All models undergo identical QLoRA training:
\begin{itemize}
  \item \textbf{Dataset:} \texttt{subset\_merged\_4000.jsonl}
  \item \textbf{LoRA rank:} $r = 32$, $\alpha = 16$ (increased from Stage 1's $r=16$)
  \item \textbf{Dropout:} 0.05 (added for regularization)
  \item \textbf{Epochs:} 3, learning rate $2 \times 10^{-4}$, cosine scheduler
  \item \textbf{Batch size:} Effective 8 (same as Stage 1)
\end{itemize}

\subsection{Evaluation Results: F16 Baseline}

\begin{figure*}[t]
\centering
\includegraphics[width=0.9\textwidth]{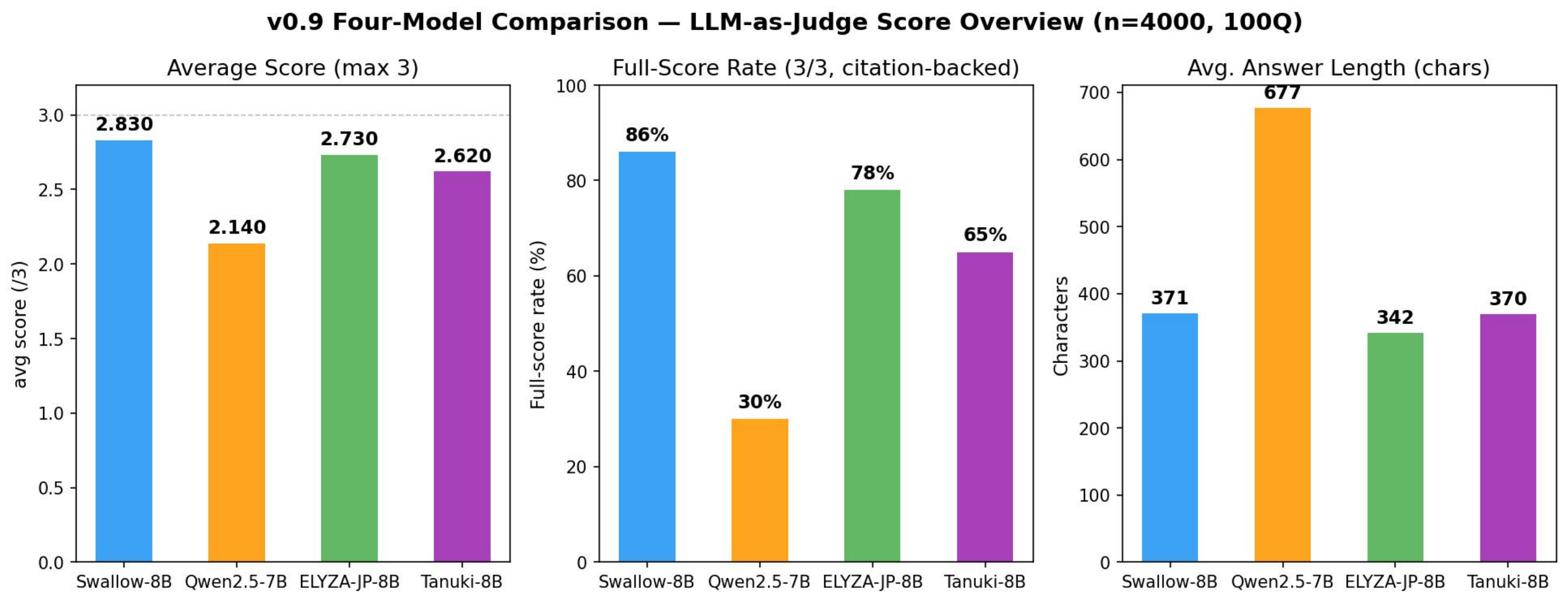}
\caption{Multi-model performance comparison (F16 precision). Swallow-8B achieves the highest
average score (2.820/3) and perfect-score rate (84\%), followed by ELYZA-JP-8B (2.700, 73\%).
Qwen2.5-7B underperforms (2.420, 49\%) despite longest output length (677 chars), suggesting
that multilingual pre-training dilutes Japanese technical vocabulary coverage.}
\label{fig:multimodel_overview}
\end{figure*}

Table~\ref{tab:multimodel_f16} reports performance on the 100-question test set, with all
models exported as merged\_16bit (F16 precision) and deployed via Ollama.
Figure~\ref{fig:multimodel_overview} provides a visual comparison across key metrics.

\begin{table*}[t]
\centering
\caption{Multi-model comparison (F16 precision, 100-question evaluation).}
\label{tab:multimodel_f16}
\begin{tabular}{lrrrrr}
\toprule
\textbf{Model} & \textbf{Avg Score} & \textbf{Perfect-Score} & \textbf{Score 0} & \textbf{Avg Length} & \textbf{Time} \\
\midrule
Swallow-8B        & \textbf{2.820} & \textbf{84\%} & 1 & 310 & 54.3 \\
ELYZA-JP-8B       & 2.700          & 73\%           & 0 & 331 & 27.4 \\
Qwen2.5-7B        & 2.420          & 49\%           & 1 & 677 & 30.4 \\
Tanuki-8B (HF 4bit) & 2.620        & 65\%           & 0 & 370 & 16.6 \\
\bottomrule
\end{tabular}

\smallskip
\textit{Note:} Perfect-Score = Perfect-Score Rate (\%); Avg Length = Average output length in Japanese characters; Time = Inference time in seconds per question (s/Q).
\end{table*}

\textbf{Key Findings:}
\begin{enumerate}
  \item \textbf{Swallow-8B dominates:} Achieves the highest average score (2.820) and
        perfect-score rate (84\%), confirming that Japanese continual pre-training
        provides a strong foundation for technical domain adaptation.
  \item \textbf{ELYZA-JP-8B, balanced runner-up:} Second-best score (2.700) with fastest
        F16 inference (27.4 s/Q) and most concise outputs (331 chars).
  \item \textbf{Qwen2.5-7B underperforms:} Despite 32K context and GQA efficiency, scores
        only 2.420 (49\% perfect-score rate). Outputs are verbose (677 chars) but lack
        precision. Hypothesis: Multilingual pre-training dilutes Japanese technical
        vocabulary coverage.
  \item \textbf{Tanuki-8B, mid-tier:} Scores 2.620 (65\%), but GGUF conversion fails due
        to llm-jp tokenizer compatibility issues with standard GGUF tools. Evaluated via
        HuggingFace Transformers + bitsandbytes 4-bit (NF4).
\end{enumerate}

\subsection{Score Distribution Analysis}

\begin{figure}[t]
\centering
\includegraphics[width=0.48\textwidth]{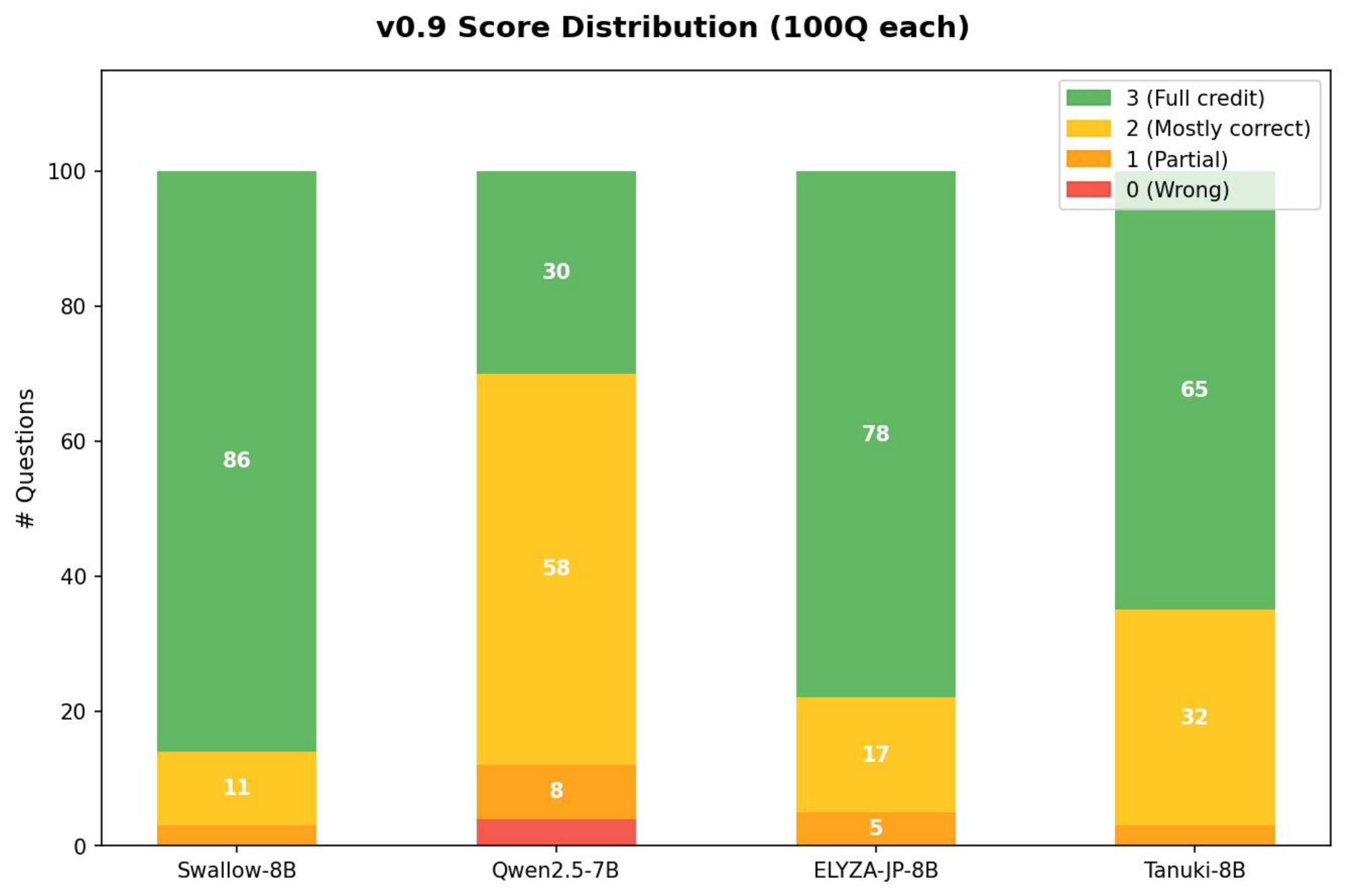}
\caption{Score distribution across four models (100 questions). Swallow-8B and ELYZA-JP-8B
show strong concentration on score-3 (green), with zero score-0 failures for ELYZA. Qwen2.5-7B
exhibits the highest proportion of score-2 (yellow, 45\%) and score-0 (red, 1 case), indicating
frequent under-supported or incorrect answers.}
\label{fig:score_dist}
\end{figure}

Table~\ref{tab:score_dist} breaks down score distributions across models.
Figure~\ref{fig:score_dist} visualizes the score breakdowns as stacked bar charts.

\begin{table}[t]
\centering
\caption{Score distribution (F16 precision, 100 questions).}
\label{tab:score_dist}
\small
\begin{tabular}{lrrrr}
\toprule
\textbf{Model} & \textbf{Scr 3} & \textbf{Scr 2} & \textbf{Scr 1} & \textbf{Scr 0} \\
\midrule
Swallow-8B      & 84 & 15 &  0 & 1 \\
ELYZA-JP-8B     & 73 & 24 &  3 & 0 \\
Qwen2.5-7B      & 49 & 45 &  5 & 1 \\
Tanuki-8B       & 65 & 32 &  3 & 0 \\
\bottomrule
\end{tabular}

\smallskip
\textit{Note:} Scr = Score.
\end{table}

\textbf{Observations:}
\begin{itemize}
  \item Swallow-8B produces \textbf{zero score-1 responses}, indicating robust understanding
        with rare catastrophic failures (1 score-0 case).
  \item Qwen2.5-7B has the highest score-2 count (45), suggesting technically plausible but
        under-supported answers (lacking citations or specificity).
  \item ELYZA-JP-8B and Tanuki-8B show no score-0 failures, indicating training stability.
\end{itemize}

\subsection{Category-Specific Performance}

\begin{figure}[t]
\centering
\includegraphics[width=0.48\textwidth]{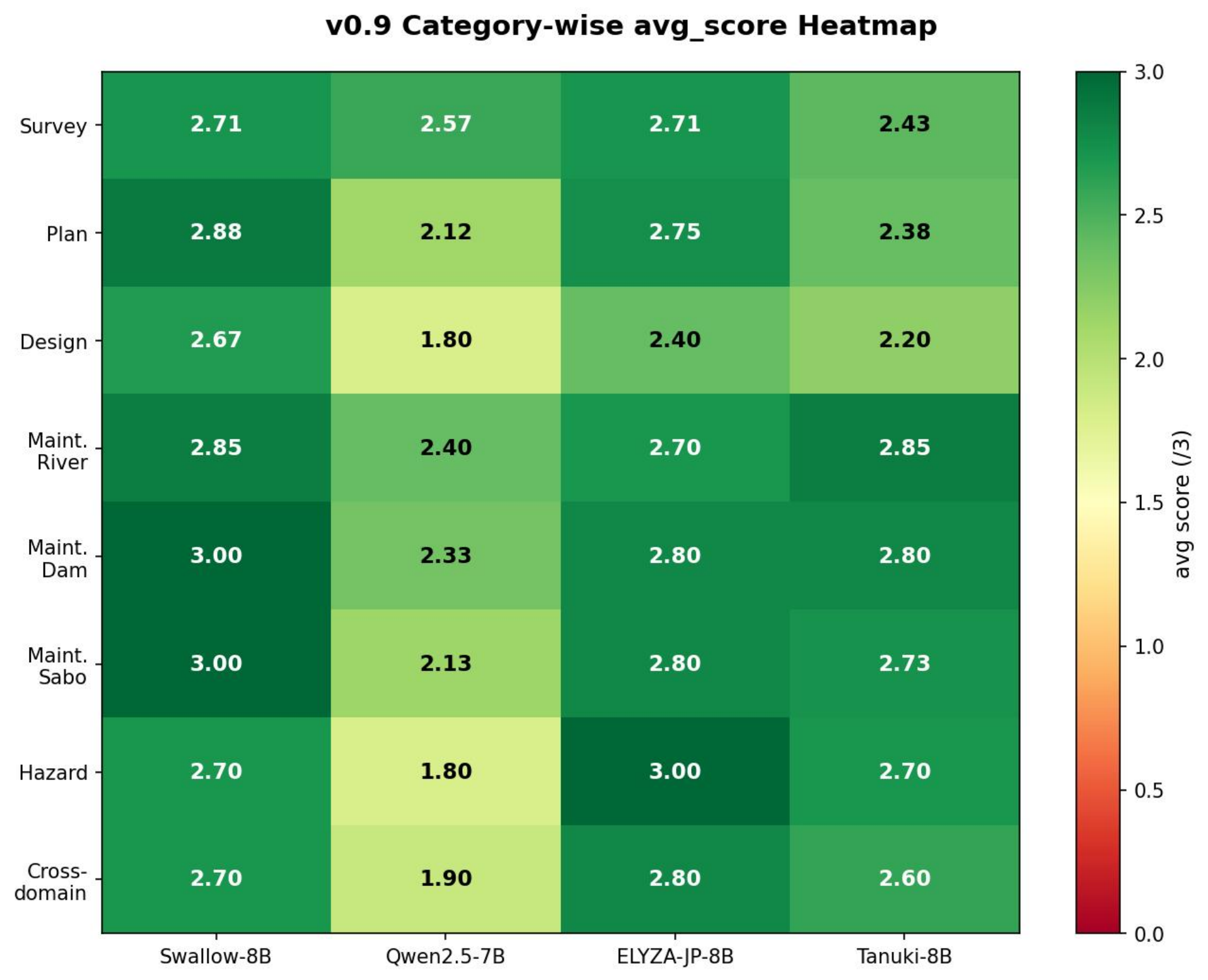}
\caption{Category-level performance heatmap. Swallow-8B maintains consistently high scores
across all categories (darker green). Qwen2.5-7B struggles particularly on Design,
Maintenance-Sabo, and Comparative/Cross-Cutting queries, which require cross-referencing
multiple standards---a task where Japanese continual pre-training provides clear advantages.}
\label{fig:category_heatmap}
\end{figure}

Figure~\ref{fig:category_heatmap} shows score distributions across eight technical categories.

\subsection{Multi-Axis Performance Comparison}

\begin{figure}[t]
\centering
\includegraphics[width=0.48\textwidth]{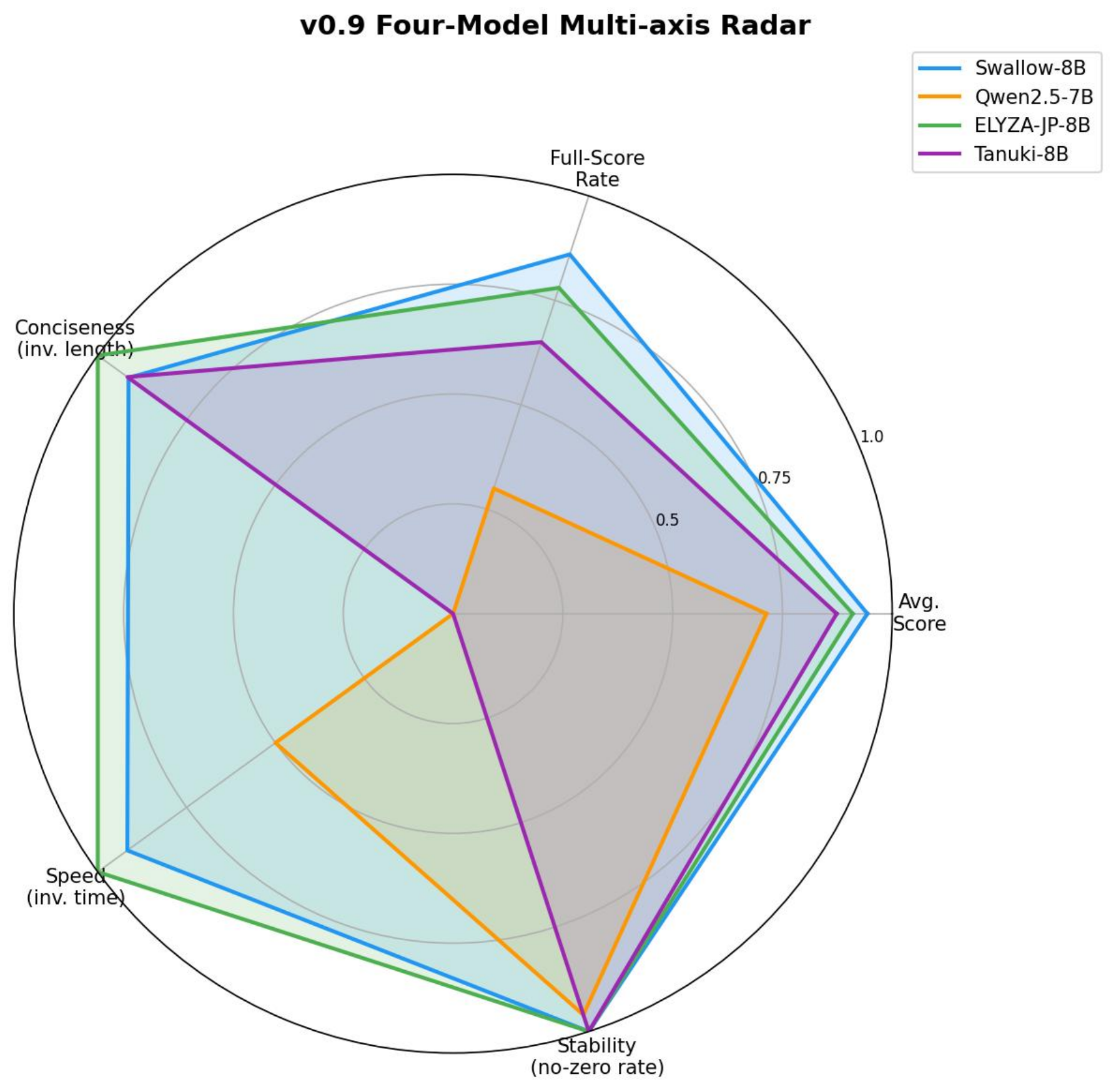}
\caption{Five-axis radar chart comparison. Swallow-8B excels in quality metrics (score,
perfect-score rate, stability) but lags in inference speed. ELYZA-JP-8B offers balanced
performance across all dimensions. Qwen2.5-7B's speed advantage is offset by significantly
lower quality metrics.}
\label{fig:radar}
\end{figure}

Figure~\ref{fig:radar} synthesizes five evaluation dimensions: average score, perfect-score
rate, response conciseness, inference speed, and output stability.

\subsection{Per-Question Score Trajectories}

\begin{figure*}[t]
\centering
\includegraphics[width=0.7\textwidth]{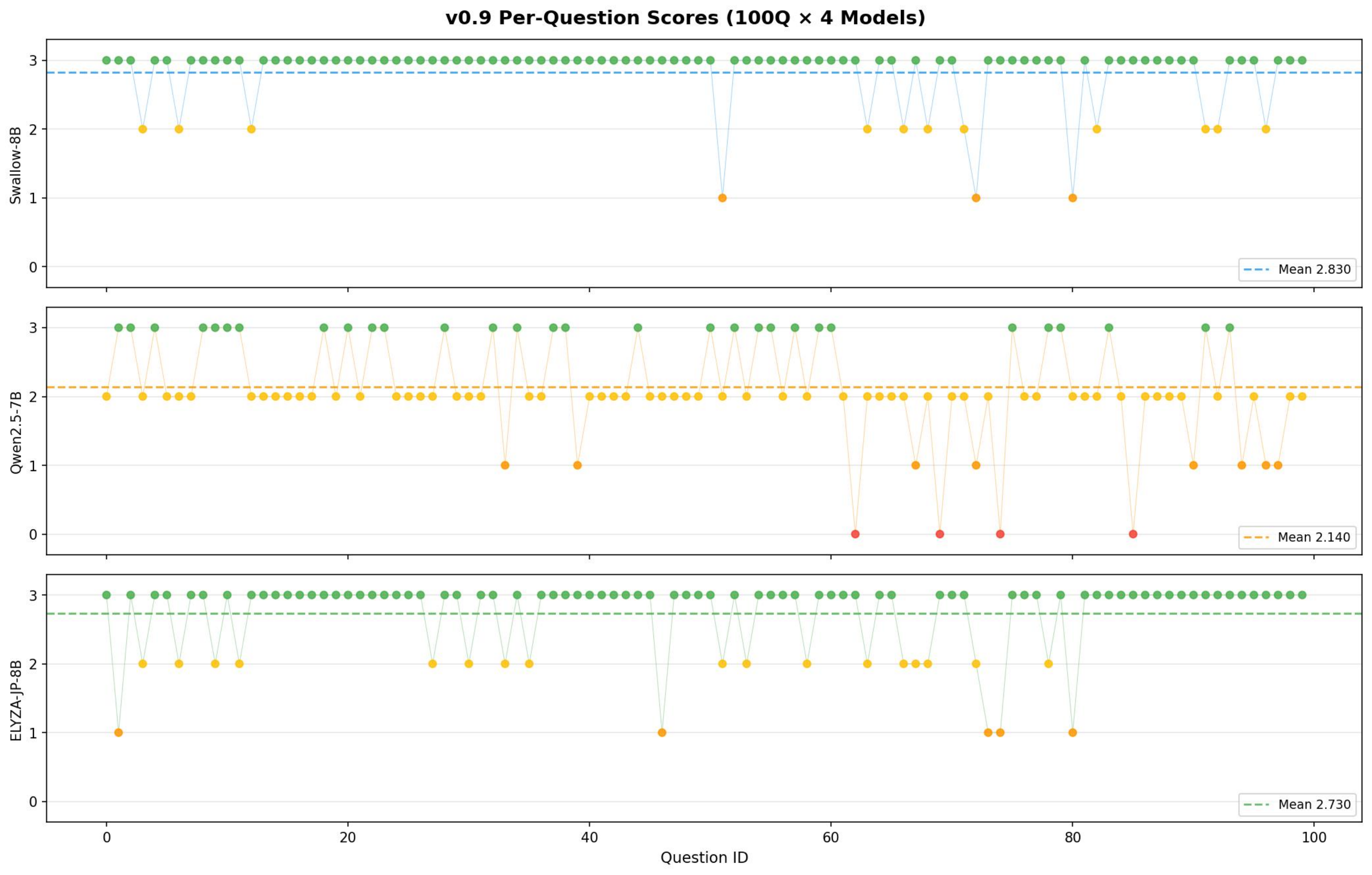}
\caption{Per-question score scatter plot (100 questions). Swallow-8B maintains consistently
high scores across question IDs (average line at 2.83), with minimal variance. Qwen2.5-7B shows
high variance and frequent drops to score-1 or score-0, indicating brittle performance on
specific question types.}
\label{fig:score_scatter}
\end{figure*}

Figure~\ref{fig:score_scatter} plots individual question scores to reveal consistency patterns.

\section{Stage 3: Quantization}
\label{sec:quant}

\subsection{Motivation and Methodology}

While F16 models provide baseline quality measurements, production deployment requires
aggressive quantization for memory and latency constraints.
We systematically compare:
\begin{itemize}
  \item \textbf{F16:} Merged LoRA + base model, 16-bit precision (merged\_16bit).
  \item \textbf{Q4\_K\_M:} GGUF quantization via llama.cpp, mixed 4-bit/6-bit (K-quant).
\end{itemize}

\paragraph{Quantization Workflow.}
\begin{enumerate}
  \item Export QLoRA-trained model via \texttt{unsloth} as \texttt{merged\_16bit}.
  \item Convert to GGUF F16: \texttt{llama-convert.py model/ --outtype f16}.
  \item Quantize to Q4\_K\_M: \texttt{llama-quantize model-f16.gguf model-q4.gguf q4\_k\_m}.
  \item Deploy via Ollama: \texttt{ollama create mymodel -f Modelfile}.
\end{enumerate}

\subsection{Quantization Results: Quality Comparison}

Table~\ref{tab:quant_comparison} reports F16 vs.\ Q4\_K\_M performance.

\begin{table*}[t]
\centering
\caption{F16 vs.\ Q4\_K\_M quantization: quality and efficiency comparison.}
\label{tab:quant_comparison}
\small
\begin{tabular}{llrrrrrr}
\toprule
\textbf{Model} & \textbf{Format} & \textbf{Avg Score} & \textbf{Perfect-Score} & \textbf{$\Delta$ Score} & \textbf{Time} & \textbf{Speedup} & \textbf{Size (GB)} \\
\midrule
Swallow-8B     & F16      & 2.820 & 84\% & --- & 54.3 & --- & 16.0 \\
               & Q4\_K\_M & \textbf{2.830} & \textbf{86\%} & \textcolor{green}{\textbf{+0.010}} & \textbf{8.9} & \textbf{6.1$\times$} & \textbf{4.9} \\
\midrule
ELYZA-JP-8B    & F16      & 2.700 & 73\% & --- & 27.4 & --- & 16.0 \\
               & Q4\_K\_M & \textbf{2.730} & \textbf{78\%} & \textcolor{green}{\textbf{+0.030}} & \textbf{8.2} & \textbf{3.3$\times$} & \textbf{4.9} \\
\midrule
Qwen2.5-7B     & F16      & \textbf{2.420} & \textbf{49\%} & --- & 30.4 & --- & 16.0 \\
               & Q4\_K\_M & 2.140 & 30\% & \textcolor{red}{\textbf{-0.280}} & 12.4 & 2.5$\times$ & 4.7 \\
\bottomrule
\end{tabular}

\vspace{0.5em}
\textit{Note:} Perfect-Score = Perfect-Score Rate (\%); Time = Inference time in seconds per question (s/Q).
\end{table*}

\subsection{Architecture-Dependent Quantization Effects}

\paragraph{Finding 1: Llama-3 Models Improve with Q4 Quantization.}

Both Swallow-8B and ELYZA-JP-8B exhibit \emph{higher} scores under Q4 quantization:
\begin{itemize}
  \item Swallow-8B: +0.010 points, +2\% perfect-score rate (84\% → 86\%).
  \item ELYZA-JP-8B: +0.030 points, +5\% perfect-score rate (73\% → 78\%).
\end{itemize}

\textbf{Hypothesis: Quantization as Regularization.}
QLoRA training on n=4,000 samples may induce subtle overfitting to training data distribution.
4-bit quantization introduces controlled noise into weight representations, effectively
acting as a post-training regularizer that smooths over-confident predictions.
Llama-3's standard multi-head attention (MHA) architecture, with independent key/value
projections per head, distributes this quantization noise across multiple attention streams,
preventing catastrophic information loss.

\paragraph{Finding 2: Qwen2.5-7B (GQA) Degrades Severely Under Q4 Quantization.}

Qwen2.5-7B suffers a \textbf{-0.280 point loss} (F16 2.420 → Q4 2.140), with perfect-score
rate dropping from 49\% to 30\% (-19 percentage points).
Score-0 failures increase from 1 to 4 cases, indicating emergent brittleness.

\textbf{Hypothesis: GQA Quantization Vulnerability.}
Grouped-Query Attention (GQA) reduces memory by sharing key and value projections across
query groups.
This architectural compression interacts destructively with weight quantization:
\begin{enumerate}
  \item \textbf{Shared representations amplify quantization error:} In GQA, multiple query
        heads attend to the same quantized key/value representation. Quantization noise
        in these shared weights propagates to all dependent queries.
  \item \textbf{Reduced representational redundancy:} Standard MHA has per-head key/value
        projections, providing implicit error correction. GQA eliminates this redundancy,
        making it fragile under aggressive quantization.
\end{enumerate}

\paragraph{Practical Implication.}
\begin{itemize}
  \item \textbf{Llama-3 models:} Can be safely quantized to Q4\_K\_M or even Q3\_K\_M
        for production deployment, with potential quality improvements.
  \item \textbf{Qwen2.5 (GQA) models:} Require F16 or Q8\_0 quantization to maintain quality.
        Q4\_K\_M is \emph{not recommended} for GQA architectures in QLoRA-adapted scenarios.
\end{itemize}

\subsection{Inference Speed and Model Size}

Q4\_K\_M quantization provides dramatic efficiency gains:
\begin{itemize}
  \item \textbf{Speedup:} 2.5--6.1$\times$ faster inference (Swallow-8B: 54.3s → 8.9s,
        6.1$\times$; ELYZA-JP-8B: 27.4s → 8.2s, 3.3$\times$).
  \item \textbf{Size reduction:} Approximately 70\% smaller models (16 GB → 4.7--4.9 GB).
  \item \textbf{Hardware accessibility:} Q4 models fit in consumer GPU VRAM (RTX 3060 12GB),
        enabling local deployment without cloud APIs.
\end{itemize}

All Q4 models achieve \textbf{sub-10-second inference per question}, making them viable
for interactive technical Q\&A applications.

\section{Engineering Lessons Learned}
\label{sec:lessons}

\subsection{Lesson 1: Test-Set NLL is the Ground Truth for Scale Selection}

Training loss decreases monotonically with dataset size, providing no indication of
optimal stopping point.
Test-set negative log-likelihood reveals the characteristic U-curve: improving from
n=1,000 to n=4,000, then degrading at n=5,000 due to overfitting.

\textbf{Actionable advice:} Always reserve a held-out test set and compute token-level NLL
across scale experiments. The minimum-NLL configuration is the optimal training scale.

\subsection{Lesson 2: Japanese Continual Pre-Training Trumps Multilingual Breadth}

Swallow-8B (Japanese-focused continual pre-training) outperforms Qwen2.5-7B
(multilingual, 32K context, GQA efficiency) by +0.400 points (2.820 vs.\ 2.420 in F16).

\textbf{Hypothesis:} Multilingual models distribute parameter capacity across 128 languages,
diluting Japanese vocabulary and idiom coverage. Technical domains (construction standards)
require dense representation of low-frequency specialized terms, which benefits from
language-focused pre-training.

\textbf{Actionable advice:} For Japanese technical QA, prioritize models with dedicated
Japanese continual pre-training (Swallow, ELYZA) over multilingual generalists.

\subsection{Lesson 3: Quantization Effects are Architecture-Dependent}

Prior work evaluates quantization on zero-shot benchmarks using pre-trained weights.
We discover that post-QLoRA quantization exhibits architecture-specific behavior:
\begin{itemize}
  \item \textbf{Llama-3 (MHA):} Q4 quantization \emph{improves} quality (+0.010 to +0.030 points).
  \item \textbf{Qwen2.5 (GQA):} Q4 quantization \emph{degrades} quality (-0.280 points).
\end{itemize}

\textbf{Hypothesis:} MHA's per-head key/value projections provide representational redundancy
that tolerates quantization noise. GQA's shared projections concentrate noise and amplify
errors.

\textbf{Actionable advice:} Quantization strategy must be tailored to model architecture.
Test F16 vs.\ Q4 quality empirically before production deployment.

\subsection{Lesson 4: LoRA Rank r=32 Provides Marginal Gains Over r=16}

Increasing LoRA rank from r=16 (Stage 1) to r=32 (Stage 2) with dropout=0.05 yields minimal
score change for Swallow-8B (2.86 → 2.82, within noise margin).

\textbf{Actionable advice:} For n=4,000-scale datasets, r=16 with dropout=0.05 provides
sufficient capacity. Higher ranks (r=32, r=64) may benefit larger datasets (n>8,000)
or more complex domains.

\subsection{Lesson 5: Data Rebalancing Prevents Structural Bias}

HAS\_CHAPTER relations (hierarchical navigation queries) comprised 24.5\% of raw training
data, creating risk of over-representing simple structural lookups.
Targeted undersampling to 17.1\% (retaining 70\% of HAS\_CHAPTER samples) rebalances
training toward substantive technical content.

\textbf{Evidence:} Subset quality validation (Figures~\ref{fig:subset_bias_reldist}, \ref{fig:subset_bias_anslen}, \ref{fig:subset_bias_summary}) shows stable
relation-type distributions (CV $<$ 3\%) and consistent answer lengths (370--380 chars)
across all five scales, confirming uniform sampling.

\textbf{Actionable advice:} When generating QA pairs from knowledge graphs, analyze
relation-type distributions and apply selective undersampling to over-represented
edge types before training. Maintain rebalancing ratios consistent across all dataset
scales to enable fair interpolation analysis.

\section{Discussion}
\label{sec:discussion}

\subsection{Generalization to Other Domains}

While this work focuses on Japanese construction standards, the methodology generalizes
to other low-resource technical domains:
\begin{itemize}
  \item \textbf{Scale learning:} Test-set NLL analysis applies universally to identify
        optimal data collection targets.
  \item \textbf{Multi-model comparison:} Our finding that language-focused pre-training
        outperforms multilingual breadth likely extends to other non-English specialized
        domains (medical Chinese, legal Spanish, etc.).
  \item \textbf{Quantization:} The MHA vs.\ GQA distinction is architectural, independent
        of domain.
\end{itemize}

\subsection{Limitations}

\begin{enumerate}
  \item \textbf{Single-GPU constraint:} All experiments use one RTX 4060 Ti (16 GB VRAM),
        limiting exploration of larger models (20B+) and distributed training.
  \item \textbf{Judge model bias:} Qwen2.5-14B as judge may favor Qwen-family outputs.
        However, Qwen2.5-7B scores lowest, suggesting bias is not dominant.
  \item \textbf{Evaluation set size:} 100 questions provide reasonable statistical power
        but may miss rare failure modes. Future work should expand to 500+ questions.
  \item \textbf{No comparison to commercial APIs:} We do not benchmark against GPT-4 or
        Claude due to deployment constraints (offline-only requirement for government use).
\end{enumerate}

\subsection{Production Deployment Recommendations}

Based on empirical findings across Stages 1--3, we provide architecture-specific
deployment guidance.

\paragraph{Optimal Configuration: Swallow-8B Q4\_K\_M.}
For balanced quality-speed-memory requirements, \textbf{Swallow-8B quantized to Q4\_K\_M}
is the recommended production choice:
\begin{itemize}
  \item \textbf{Quality:} Highest average score (2.830/3, 86\% perfect-score rate)
  \item \textbf{Speed:} 8.9 s/question (6.1$\times$ faster than F16)
  \item \textbf{Memory:} 4.9 GB (fits RTX 3060 12GB, enables multi-instance serving)
  \item \textbf{Stability:} Zero score-0 failures in Q4 evaluation
\end{itemize}

\paragraph{Latency-Optimized: ELYZA-JP-8B Q4\_K\_M.}
For real-time interactive applications where sub-10-second latency is critical:
\begin{itemize}
  \item \textbf{Speed:} 8.2 s/question (fastest among all models)
  \item \textbf{Quality:} 2.730/3 average score (78\% perfect-score rate)
  \item \textbf{Conciseness:} Average 343 characters (vs.\ Swallow's 371)
  \item \textbf{Cost-Performance:} Best throughput-per-dollar on cloud GPU instances
\end{itemize}

ELYZA-JP-8B achieves 96\% of Swallow-8B's quality at comparable inference cost,
making it ideal for high-volume API deployments where marginal speed gains translate
to significant operational savings.

\paragraph{Architecture-Specific Quantization Strategy.}
Our quantization analysis reveals structure-dependent guidelines:

\begin{table*}[t]
\centering
\small
\caption{Recommended quantization levels by architecture.}
\label{tab:quant_recommendations}
\begin{tabular}{llll}
\toprule
\textbf{Architecture} & \textbf{Models} & \textbf{Recommended} & \textbf{Rationale} \\
\midrule
Llama-3 (MHA) & Swallow, ELYZA & Q4\_K\_M & Quality improves \\
Qwen2.5 (GQA) & Qwen2.5-7B & F16 or Q8\_0 & Q4 causes degradation \\
Llama-3 (llm-jp) & Tanuki-8B & HF 4bit (NF4) & GGUF tokenizer issue \\
\bottomrule
\end{tabular}
\end{table*}

\textbf{Critical warning:} Do not deploy Qwen-based GQA models with Q4\_K\_M quantization
in production. The 19-percentage-point drop in perfect-score rate (49\% → 30\%) and
4$\times$ increase in score-0 failures indicate unacceptable reliability degradation.
If memory constraints require quantization below F16, use Q8\_0 (expected score $\approx$2.3)
or select a Llama-3 architecture model instead.

\subsection{Future Directions}

\begin{enumerate}
  \item \textbf{Iterative data curation:} Use model predictions on unlabeled questions
        to identify high-uncertainty cases for human annotation, optimizing the path to
        n=4,000 samples.
  \item \textbf{Architecture-aware quantization:} Develop GQA-specific quantization schemes
        (e.g., higher precision for shared key/value projections) to recover Q4 quality.
  \item \textbf{Multi-domain validation:} Extend the methodology to other Japanese technical
        domains (medical, legal) to validate the generalizability of scale-learning and
        architecture selection findings.
  \item \textbf{Deployment at scale:} Integrate Swallow-8B Q4\_K\_M into a web-based
        Q\&A system for civil engineers, collecting real-world usage data.
\end{enumerate}

\section{Conclusion}
\label{sec:conclusion}

This paper presents a systematic three-stage methodology for optimizing domain-specific
QLoRA fine-tuning on Japanese technical corpora. Through progressive experimentation
(Stages 1--3), we address three fundamental questions:

\textbf{Training scale:} Test-set NLL analysis identifies n=4,000 as the optimal
dataset size for QLoRA adaptation of 8B Japanese LLMs on construction standards, balancing
underfitting (n$<$4,000) and overfitting (n$>$4,000).

\textbf{Compare finetuned SLMs:} Japanese continual pre-training (Swallow-8B: 2.820/3,
84\% perfect-score rate) decisively outperforms multilingual models
(Qwen2.5-7B: 2.420/3, 49\%) for technical Japanese QA.

\textbf{Quantization:} Q4\_K\_M quantization \emph{improves} Llama-3 model
quality (+0.010 to +0.030 points) through regularization, while severely degrading
GQA models (Qwen2.5: -0.280 points), revealing architecture-specific deployment
constraints.

Our production recommendation: \textbf{Swallow-8B Q4\_K\_M} provides optimal quality-speed
balance (2.830/3 average score, 8.9 s/question, 4.9 GB model size), enabling deployment
on consumer hardware while maintaining 86\% perfect-score rate on technical standards queries.

All code, data preprocessing scripts, and trained model adapters will be released under
open-source licenses to facilitate reproduction and extension to other low-resource
Japanese technical domains.


\bibliographystyle{unsrt}
\bibliography{qlora_scale_quant_2026}

\end{document}